\pdfoutput=1

\documentclass[11pt]{article}

\usepackage[final]{acl}

\usepackage{times}
\usepackage{latexsym}
\usepackage{enumitem}
\usepackage{amssymb}
\usepackage{amsmath}
\usepackage[T1]{fontenc}
\usepackage{booktabs}
\usepackage{multirow}
\usepackage{graphicx}
\usepackage{float}
\usepackage{algorithm}
\usepackage{algpseudocode}
\usepackage{booktabs}
\usepackage{xcolor}
\usepackage{colortbl} 
\usepackage{xcolor}
\usepackage{booktabs}
\usepackage{hyperref}

\usepackage[utf8]{inputenc}

\usepackage{microtype}

\usepackage{inconsolata}
\usepackage{stfloats}
\usepackage{graphicx}

\title{CALM: A CKA-Guided Adaptive Layer-Wise Modularization Framework for LLM Quantization}

\author{
    \textbf{Jinhao Zhang}$^\spadesuit$, 
    \textbf{Yunquan Zhang}$^\clubsuit$, 
    \textbf{Daning Cheng}$^\clubsuit$, 
    \textbf{Jun Sun}$^\heartsuit$, 
    \textbf{Zicheng Yan}$^\diamondsuit$ \\
    \\[0.5ex] 
    $^\spadesuit$Beijing University of Posts and Telecommunications, Beijing, China \\
    $^\clubsuit$Institute of Computing Technology, Chinese Academy of Sciences, Beijing, China \\
    $^\heartsuit$Zhejiang University, Hangzhou, China \\
    $^\diamondsuit$University of Science and Technology of China, Hefei, China
}

\begin{document}
\maketitle
\begin{abstract}

Current mainstream post-training quantization methods for large language models typically apply a uniform quantization strategy across all network layers, overlooking the substantial differences in algorithmic suitability among layers. To address this limitation, we propose CALM (A CKA-guided Adaptive Layer-wise Modularization)—a fine-tuning-free, plug-and-play framework for algorithmic heterogeneous quantization. CALM independently evaluates multiple PTQ algorithms on each layer and employs Linear Centered Kernel Alignment (CKA) as a metric to automatically select the optimal quantization strategy per layer. The individually optimized strategies are then integrated to construct a hybrid quantized model. Experiments demonstrate that our approach consistently outperforms both uniform quantization baselines and state-of-the-art mixed-precision methods across mainstream LLMs—including LLaMA and Qwen—in terms of perplexity (PPL) and downstream task performance.
\end{abstract}

\section{Introduction}

Post-Training Quantization (PTQ) has emerged as a cornerstone for compressing and deploying Large Language Models (LLMs). Existing PTQ methods—such as GPTQ, AWQ, and SmoothQuant—typically enforce a uniform quantization strategy across all network layers. While this simplifies implementation, it overlooks an important empirical fact: distinct transformer layers exhibit vastly different sensitivities to quantization errors and outlier distributions.

This variation arises from fundamental differences in how these methods operate. For instance, GPTQ reduces quantization error by optimizing weights after quantization; AWQ preserves salient weights to maintain the activation distribution; and SmoothQuant improves stability at low bit-widths by smoothing activations before quantization. These complementary strategies suggest that no single method is optimal for every layer.

Motivated by this observation, we ask:  
\textbf{Can we construct a single quantized model by selecting the most effective quantization method for each layer—combining them to achieve better overall performance than any uniform approach?}

\begin{figure}[ht]
    \centering
    \includegraphics[width=0.5\textwidth]{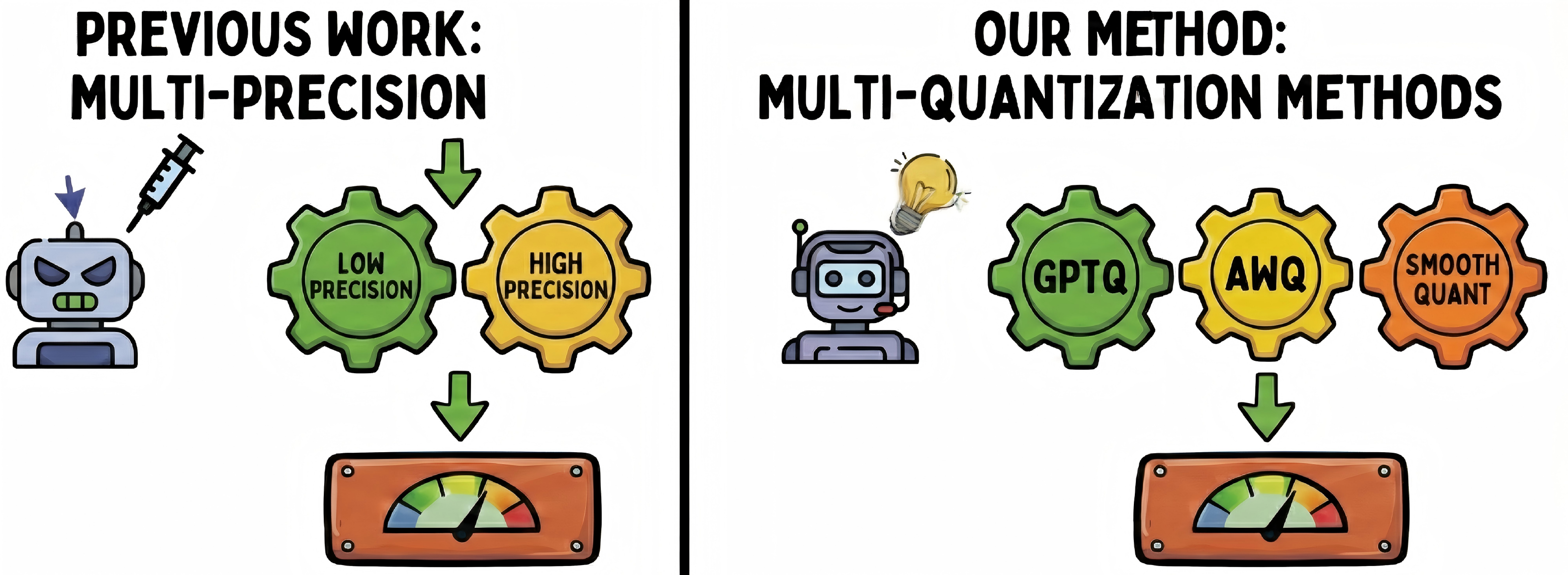} 
    \caption{Comparison between previous multi-precision quantization and our proposed multi-quantization-methods approach. Left: existing methods use different bit-widths (e.g., low vs. high precision) but a single quantization algorithm across layers. Right: our method applies diverse quantization algorithms (e.g., GPTQ, AWQ, SmoothQuant) to different layers, enabling algorithmic heterogeneity for improved performance.}
    \label{carton}
\end{figure}

Our approach differs from conventional mixed-precision quantization (e.g., using INT4 in some layers and INT8 in others), which varies only the bit-width while keeping the quantization algorithm fixed. In contrast, we explore {algorithmic heterogeneity}: applying different quantization algorithms to varying layers to better match their individual characteristics, shown in Figure \ref{carton}.

To implement this idea, we propose CKA-guided Adaptive Layer-wise Modularization (CALM). We use Linear Centered Kernel Alignment (CKA)—a well-established metric for measuring similarity between neural network representations—as a proxy for functional fidelity. Specifically, for each decoder layer, we apply multiple PTQ methods independently and compute the CKA similarity between the activations of each quantized variant and those of the original full-precision model, using a small calibration dataset. The method yielding the highest CKA score is selected for that layer, and the chosen layers are assembled into a single heterogeneous quantized model.

We evaluate CALM on several LLMs, including LLaMA and Qwen, across language modeling and downstream tasks. Results show that CALM consistently outperforms both individual PTQ baselines and state-of-the-art mixed-precision methods—without any retraining or fine-tuning.

Our work demonstrates that achieving an optimal trade-off between efficiency and accuracy in LLM quantization requires not only choosing appropriate bit-widths but also selecting the right quantization algorithm for each layer. The proposed framework is fully post-training, training-free, and compatible with any off-the-shelf PTQ method, making it a practical plug-and-play enhancement for existing compression pipelines.

In summary, our key contributions are:
\begin{itemize}[leftmargin=1em]
    \item We present the first systematic study of combining diverse PTQ methods at the layer level and propose a general framework for optimal integration.
    \item We introduce a layer-wise quantization method selection strategy based on Linear CKA, enabling data-driven per-layer algorithm choice.
    \item Extensive experiments show that CALM achieves state-of-the-art performance across multiple models and benchmarks.
\end{itemize}

\section{Related work}
\textbf{Post-Training Quantization} Foundational works such as AdaRound \citep{DBLP:conf/icml/NagelABLB20}, GPTQ \citep{GPTQ}, AWQ \citep{AWQ}, and SmoothQuant \citep{SmoothQuant} have established widely adopted paradigms for minimizing quantization error through weight reconstruction and outlier mitigation. Building on these baselines, recent advanced methods introduce novel perspectives: QuIP \citep{DBLP:conf/nips/CheeCKS23} utilizes incoherence processing to enhance extreme low-bit quantization, while OmniQuant \citep{DBLP:conf/iclr/ShaoC0XZLZ00024} introduces learnable clipping thresholds. Furthermore, SpinQuant \citep{spinquant} optimizes rotation matrices to suppress outliers before quantization.

\textbf{Centered Kernel Alignment} Centered Kernel Alignment (CKA), particularly its linear variant \citep{CKA,DBLP:conf/nips/MorcosRB18}, has emerged as a standard proxy for gauging neural representational similarity. It has been widely employed to analyze training dynamics \citep{DBLP:conf/iclr/NguyenRK21}, compare diverse architectures \citep{DBLP:conf/nips/RaghuUKZD21}, and guide model compression \cite{DBLP:conf/iccv/TungM19,DBLP:conf/nips/MaFW23}. Notably, high CKA similarity is often interpreted as a signal of redundancy, utilized for expert pruning in Mixture-of-Experts models \citep{MOE_QUANT} and model merging in KnOTS \citep{modelmerge}.


\section{Methodology}
\begin{figure*}[ht]
    \centering
    \includegraphics[width=0.7\textwidth]{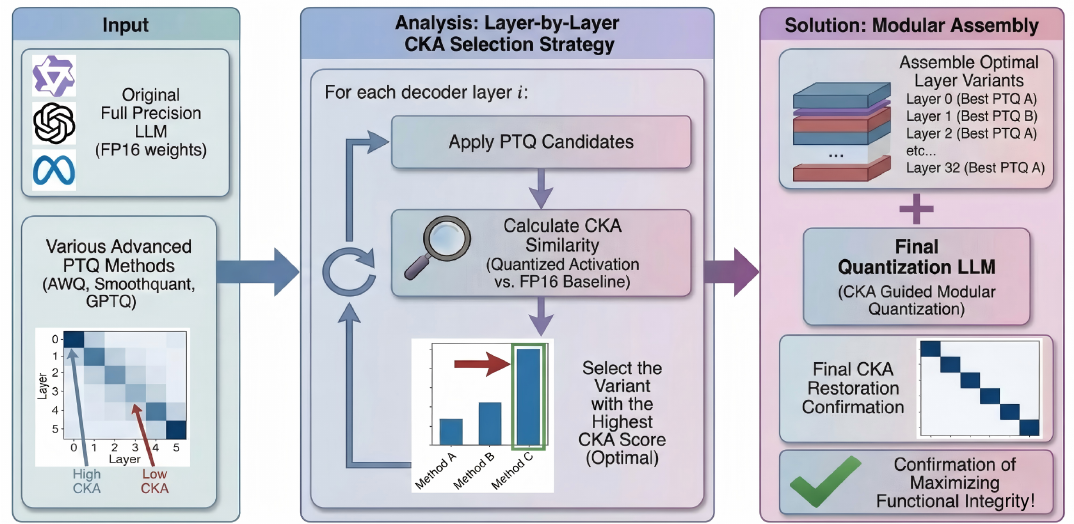} 
    \caption{\textbf{Overview of the CALM framework} (a) We first analyze layer-wise sensitivity using CKA. (b) Then, we competitively select the optimal quantization method (e.g., GPTQ \& SmoothQuant) for each layer. (c) Finally, we integrate these layers into a unified model. This framework achieves optimal heterogeneity at the algorithmic level without retraining.}
    \label{main_architecture}
\end{figure*}
\begin{figure}[ht]
    \centering
    \includegraphics[width=0.5\textwidth]{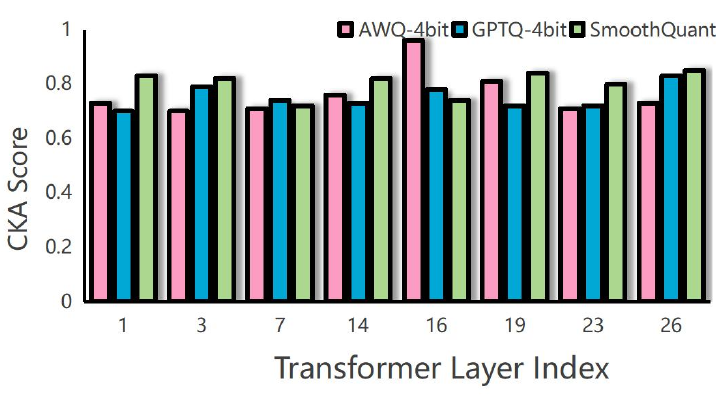} 
    \caption{Layer-wise CKA score distribution of Llama-3-8B under different quantization methods.}
    \label{CKA_result}
\end{figure}
\subsection{Algorithmic Suitability}
State-of-the-art PTQ algorithms are built upon distinct and complementary design principles, leading to varying suitability across layers. GPTQ employs a layer-wise greedy optimization approach, formulating quantization error minimization as a constrained quadratic programming problem and updating weights sequentially using Hessian information; it is particularly effective for layers with concentrated weight distributions and few outliers. In contrast, AWQ identifies “sensitive” weight channels—those most influential to the output—based on activation magnitudes and preserves higher precision for them while aggressively quantizing the rest. This method assumes a strong correlation between weight importance and the magnitude of corresponding activations, enabling superior performance when activation distributions are skewed and contain prominent hotspot channels. SmoothQuant, on the other hand, introduces a smoothing factor between weights and activations to shift quantization difficulty from the activation side to the weight side. By reparameterizing the forward pass to balance dynamic ranges, it demonstrates greater robustness in scenarios involving extreme weight outliers and large variations in activation magnitudes.

Clearly, none of these algorithms is universally optimal across all network layers: GPTQ favors smooth and compact weight distributions, AWQ relies on activation-guided sparse importance structures, and SmoothQuant excels at handling high-dynamic-range weights and activations. Consequently, the statistical characteristics of each layer’s weights—such as kurtosis, skewness, and outlier ratio—directly determine which quantization strategy is most effective. A single algorithm is unlikely to achieve optimal performance across an entire model, highlighting the necessity of layer-adaptive quantization.

Centered Kernel Alignment (CKA) is a widely used metric for measuring the similarity between two feature representation spaces and is commonly applied in the evaluation of quantization strategies. We use CKA to compare the decoder output features of LLaMA-8B models quantized by different PTQ methods against those of the full-precision model, thereby assessing how effectively each strategy preserves the original representational fidelity. As shown in the 
Figure \ref{CKA_result}, no single quantization method consistently maintains optimal performance across all layers. This variation reveals the limitations of monolithic, one-size-fits-all quantization approaches
\subsection{The CALM Framework}

Motivated by the observed layer-wise heterogeneity in algorithmic suitability, we propose CALM, whose overall pipeline is illustrated in Algorithm~\ref{alg:cka_guided} and Figure~\ref{main_architecture}. The core objective of this framework is to allow each layer to select the quantization method that best preserves its representational structure, rather than enforcing a single homogeneous strategy across the entire model. The proposed framework consists of three main stages. 

\begin{algorithm}[t]
\caption{CALM}
\label{alg:cka_guided}
\small 
\begin{algorithmic}[1]
\Require Pre-trained Full-Precision LLM layers $\mathcal{M}_{FP}=\{l_{0}, \dots, l_{L-1}\}$; Calibration Dataset $\mathcal{D}_{cal}$; 
Candidate PTQ Pool $\mathcal{P} = \{\text{GPTQ}, \text{AWQ}, \text{SmoothQuant}, \dots\}$;
Target Bit-width $B$.
\Ensure Heterogeneously Quantized Model $\mathcal{M}_{Hybrid}$.
\State Initialize $\mathcal{M}_{Hybrid} \leftarrow \emptyset$
\State Extract initial input features $X_{ref}$ from $\mathcal{D}_{cal}$ \Comment{Reference FP input}
\State Initialize quantized stream input $H_{in} \leftarrow X_{ref}$ \Comment{Input from prev quantized layer}
\For{$l = 0$ to $L-1$}
    \State \textbf{Step 1: Compute Ground Truth}
    \State Get FP weights $W_{l}$ of layer $l$
    \State Compute target activation: $H_{target}^{(l)} \leftarrow l_l(X_{ref})$
    \State \textbf{Step 2: Search Optimal Method}
    \State $S_{best} \leftarrow -1$, $l_{best} \leftarrow \text{Null}$, $H_{out}^{best} \leftarrow \text{Null}$
    \For{each method $m \in \mathcal{P}$}
        \State Quantize layer: $\hat{l}^m \leftarrow \text{Quantize}(l_l, m, B)$
        \State Forward with previous quantized input: 
        \State \quad $H_{m}^{(l)} \leftarrow \hat{l}^m(H_{in})$ \Comment{Matches Eq.(1) $f_m(H_{in})$}
        \State Calculate Fidelity:
        \State \quad $S_{CKA} \leftarrow \text{CKA}(H_{target}^{(l)}, H_{m}^{(l)})$
        
        \If{$S_{CKA} > S_{best}$}
            \State $S_{best} \leftarrow S_{CKA}$
            \State $l_{best} \leftarrow \hat{l}^m$
            \State $H_{out}^{best} \leftarrow H_{m}^{(l)}$
        \EndIf
    \EndFor
    \State \textbf{Step 3: Update \& Assemble} 
    \State $\mathcal{M}_{Hybrid}.\text{append}(l_{best})$
    \State Update inputs for next layer:
    \State \quad $H_{in} \leftarrow H_{out}^{best}$ \Comment{Pass quantized output to next layer}
    \State \quad $X_{ref} \leftarrow H_{target}^{(l)}$ \Comment{Update reference baseline}
\EndFor

\State \Return $\mathcal{M}_{Hybrid}$
\end{algorithmic}
\end{algorithm}

First, we construct a candidate pool composed of multiple Post-Training Quantization (PTQ) algorithms, such as GPTQ, AWQ, and SmoothQuant. While SpinQuant shows high accuracy, we exclude it from the candidate pool to avoid its rotation-based inference overhead, ensuring zero additional latency.

Second, we decouple an $L$-layer LLM into $L$ independent modules. For each layer $l$, all quantization methods in the candidate pool are individually applied, producing multiple quantized candidates for that layer.

Third, we compute the CKA similarity between the output activations of each quantized candidate and the corresponding full-precision activations. The mathematical definition and theoretical properties of CKA are provided in Appendix \ref{CKA}. For each layer $l$, the quantization method achieving the highest CKA score is selected as the optimal match, thereby maximizing the preservation of the layer’s intrinsic feature geometry.

\subsection{Greedy Layer-Wise Selection Strategy}

Finding the globally optimal assignment of quantization methods across all layers requires solving a combinatorial optimization problem with a search space of $|\mathcal{P}|^{L}$, which is computationally infeasible for modern large language models. To address this challenge, we adopt a greedy layer-wise selection strategy that optimizes each layer locally. While this greedy approach optimizes locally, extensive experiments confirm it is sufficient to transcend the performance ceilings of monolithic methods. CALM consistently outperforms state-of-the-art (SOTA) uniform quantization baselines across various LLMs, including LLaMA and Qwen."

This design is motivated by the sequential nature of Transformer inference. By preserving the representational structure of each layer as faithfully as possible to its full-precision counterpart, the framework provides more stable input distributions for subsequent layers, thereby mitigating the accumulation of quantization errors in deep networks.

Formally, the quantization method selected for the $l$-th layer is defined as:
\begin{equation}
m_l^{*} =
\arg\max_{m \in \mathcal{P}}
\mathrm{CKA}\!\left(
\mathbf{H}_{FP}^{(l)},
f_m\!\left(\mathbf{H}_{Q}^{(l-1)}; \mathbf{W}_l\right)
\right),
\end{equation}
where $\mathbf{H}_{FP}^{(l)}$ denotes the full-precision target activations of layer $l$, and $f_m(\cdot;\mathbf{W}_l)$ represents the forward function of layer $l$ quantized using method $m$. By conditioning on the output of the preceding quantized layer $\mathbf{H}_{Q}^{(l-1)}$, this formulation explicitly accounts for perturbations introduced by earlier layers, leading to more stable and robust layer-wise decisions in practice.

\subsection{Complexity and Inference Efficiency}

\textbf{Offline Search Cost.}
The proposed greedy search strategy exhibits linear complexity $O(L \times |\mathcal{P}|)$. For a typical 70B model with $L=80$ layers and a candidate pool of four quantization methods, only $80 \times 4 = 320$ layer-wise evaluations are required. This constitutes a one-time offline cost that is negligible compared to the expense of model pretraining or fine-tuning.

\textbf{Zero Overhead}
CALM introduces no additional latency. By finalizing layers as standard W4A8 modules with fixed group sizes, we eliminate dynamic kernel switching. Benchmarks on NVIDIA A800 GPUs confirm negligible latency deviation (~0.60\%, see Appendix \ref{app:overhead}), matching the throughput of uniform quantization.

\section{Experiments}
Our experiments are divided into three parts: PPL evaluation, downstream task evaluation, and configuration studies. The PPL and downstream experiments are designed to demonstrate the superiority of our quantized models, while the configuration experiments primarily illustrate how algorithm performance varies under different settings. These configuration studies also serve as ablation studies, helping to isolate and analyze the impact of individual design choices.

\subsection{Experiments configuration}
\textbf{Model} We evaluate the proposed method on two representative model series: Llama-3-8B, Llama-3.2-1B, Llama-3.2-3B\citep{touvron2023llama}, Qwen2.5-1.5B, and Qwen2.5-0.5B\citep{bai2023qwen}. 

\textbf{Dataset}The language modeling tasks include C4 and Wikitext2(wiki2), while the downstream tasks cover mathematical reasoning (GSM8K), code generation (HumanEval), commonsense reasoning (HellaSwag), and multi-task understanding (MMLU).

\textbf{Quantization Precision}We have quantized all models to Int4 (W4A8).

\textbf{BenchMark} We benchmark CALM against SOTA methods, including GPTQ \citep{GPTQ}, AWQ \citep{AWQ}, SmoothQuant \citep{SmoothQuant}, and SpinQuant \citep{spinquant}.

\textbf{Algorithm Configuration}
We calibrate using 128 randomly sampled sequences from the C4 dataset with a length of 1024 tokens.


\subsection{CKA-Guided Modular Quantization Results}
\begin{figure*}[ht]
    \centering
    \includegraphics[width=0.8\textwidth]{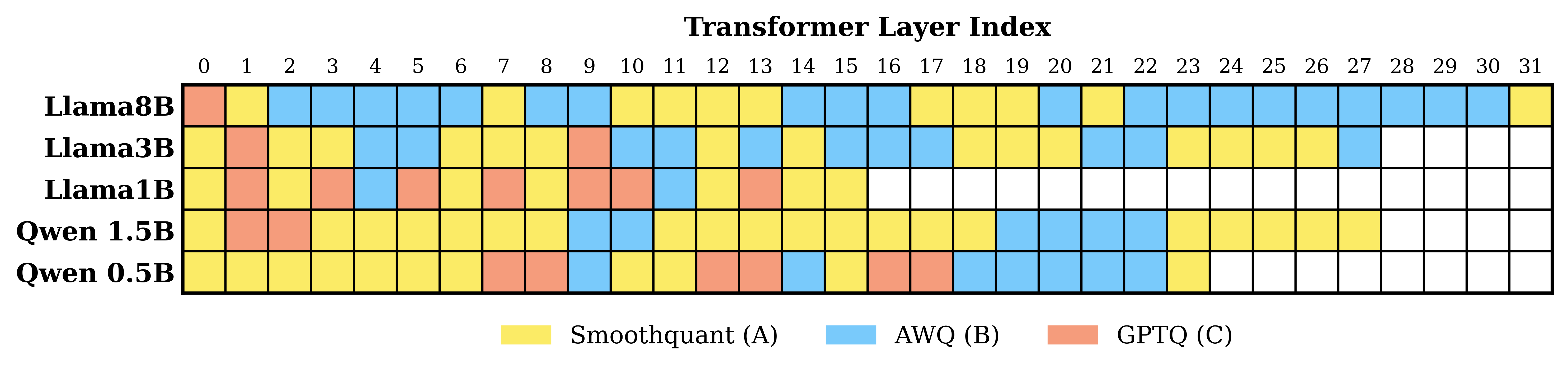} 
    \caption{Layer-wise method selection results derived from the CALM framework.}
    \label{Distribution}
\end{figure*}




Figure~\ref{Distribution} illustrates the distribution of optimal quantization methods automatically selected by the CALM framework for each layer across large language models of varying scales, including Llama8B,Llama3B, Llama1B, Qwen-0.5B, and Qwen-1.5B. It is evident from the figure that, in every model, no single quantization method dominates across the entire network depth. Instead, SmoothQuant (yellow), AWQ (blue), and GPTQ (orange) exhibit a highly interleaved and dynamically shifting allocation pattern.

Taking Llama8B as an example, its shallow layers tend to favor GPTQ, while middle to deeper layers alternate primarily between AWQ and SmoothQuant—reflecting significant differences in sensitivity to weight quantization error versus activation outliers across depths. Similarly, Llama3B employs GPTQ or SmoothQuant in its initial layers before gradually transitioning to an AWQ-dominated configuration; the smaller Llama1B, in contrast, displays a more fragmented switching behavior, indicating that even in compact models, per-layer quantization requirements remain highly heterogeneous.

The Qwen series also reveals complex patterns: Qwen-0.5B follows a staged progression of ``GPTQ $\rightarrow$ SmoothQuant $\rightarrow$ AWQ $\rightarrow$ SmoothQuant'', whereas Qwen-1.5B strongly prefers SmoothQuant in its first ten-plus layers before progressively incorporating AWQ and GPTQ. This diverse, cross-model and cross-depth selection strategy strongly suggests that different layers within large language models inherently differ in their weight distribution characteristics—such as outlier ratio, kurtosis, and skewness—as well as in their activation behaviors, leading to distinct algorithmic suitability for quantization.

\subsection{Perplexity Evaluation on LLMs}
\begin{table*}[ht]
\centering
\scriptsize 
\renewcommand{\arraystretch}{1.25} 
\setlength{\tabcolsep}{4pt} 
\begin{tabular}{lcccccccccc} 
\toprule
\multirow{2}{*}{\textbf{Method}} & \multicolumn{2}{c}{\cellcolor{gray!10}\textbf{Llama-3-8B}} & \multicolumn{2}{c}{\cellcolor{gray!10}\textbf{Llama-3.2-1B}} & \multicolumn{2}{c}{\cellcolor{gray!10}\textbf{Llama-3.2-3B}} & \multicolumn{2}{c}{\cellcolor{gray!10}\textbf{Qwen1.5-1.5B}} & \multicolumn{2}{c}{\cellcolor{gray!10}\textbf{Qwen1.5-0.5B}} \\
\cmidrule(lr){2-3} \cmidrule(lr){4-5} \cmidrule(lr){6-7} \cmidrule(lr){8-9} \cmidrule(lr){10-11}
 & \textbf{C4} $\downarrow$ & \textbf{Wiki2} $\downarrow$ & \textbf{C4} $\downarrow$ & \textbf{Wiki2} $\downarrow$ & \textbf{C4} $\downarrow$ & \textbf{Wiki2} $\downarrow$ & \textbf{C4} $\downarrow$ & \textbf{Wiki2} $\downarrow$ & \textbf{C4} $\downarrow$ & \textbf{Wiki2} $\downarrow$ \\
\midrule
FP16        & 12.28 & 6.41 & 22.48 & 11.41 & 17.24 & 9.53 & 15.95 & 10.50 & 22.34 & 13.52 \\
\midrule
AWQ         & 13.56 & 8.12 & 23.97 & 12.23 & 18.89 & 9.91 & 17.26 & 14.05 & 25.98 & 16.21 \\
GPTQ        & 14.12 & 8.81 & 24.07 & 12.18 & 18.99 & 9.82 & 17.54 & 14.50 & 26.04 & 17.03 \\
SmoothQuant & 13.64 & 7.32 & 24.01 & 12.10 & 18.87 & 9.80 & 17.33 & 13.68 & 25.34 & 15.67 \\
SpinQuant   & 13.39 & 7.48 & 23.56 & 11.89 & 18.77 & 9.75 & 16.97 & 13.31 & 25.57 & 15.34 \\
\midrule
\rowcolor{blue!10} 
\textbf{CALM} & \textbf{12.72} & \textbf{6.89} & \textbf{22.88} & \textbf{11.70} & \textbf{17.41} & \textbf{9.63} & \textbf{16.59} & \textbf{12.87} & \textbf{24.87} & \textbf{14.87} \\
\bottomrule
\end{tabular}
\caption{Perplexity (PPL) evaluation. We compare CALM with FP16 baseline and SOTA PTQ methods. }
\label{ppl_results}
\end{table*}

We evaluated the perplexity (the lower the better) of different quantization methods on multiple LLMs. The methods examined include FP16, AWQ, GPTQ, SmoothQuant, SpinQuant, and our proposed approach (CALM). We utilize the C4 and WikiText-2(Wiki2) to evaluate performance on web text and high-quality articles, respectively.

Table~\ref{ppl_results} compares the perplexity (PPL) of different quantization methods across multiple LLMs. Overall, all quantization methods lead to an increase in PPL compared to the FP16 baseline; however, the extent of this impact is contingent upon the specific model size, architecture, and quantization method employed.

Model size exhibits a significant influence on quantization robustness. For larger models (e.g., Llama-3-8B), the PPL increase on both C4 and WikiText-2 is relatively modest across all methods, including CALM, indicating that larger models are more tolerant to quantization-induced errors. In contrast, smaller models (e.g., Llama-3.2-1B and Qwen1.5-0.5B) show a more pronounced rise in PPL, suggesting greater sensitivity to quantization.

CALM achieves the best or nearly the best PPL performance across all experimental settings. Specifically:

On Llama-3-8B, CALM attains a C4 PPL of 12.72, notably lower than other quantization methods and closest to the FP16 baseline (12.28).

On Qwen1.5-1.5B, CALM reaches a WikiText-2 PPL of 12.87, which is 0.44 lower than the next best quantization method, SpinQuant (13.31).

Even on the very lightweight Qwen1.5-0.5B, CALM maintains a relatively low PPL increase (24.87 on C4), substantially outperforming AWQ (25.98) and GPTQ (26.04).

The experimental results demonstrate that CALM achieves method heterogeneity by adaptively and collaboratively selecting different quantization methods across layers. This optimal method selection, grounded in representational similarity, enables the model to surpass the performance ceiling of any single quantization method. This improvement translates directly into significant performance advantages on language modeling tasks, yielding lower perplexity degradation compared to baseline methods. 

\subsection{Downstream experiments}
To comprehensively evaluate the capabilities of quantized models in real-world tasks, we conduct a systematic assessment across multiple representative downstream tasks, covering mathematical reasoning (GSM8K), code generation (HumanEval), commonsense reasoning (HellaSwag), and multi-task understanding (MMLU). Each task follows widely accepted evaluation protocols: MMLU (5-shot), GSM8K (8-shot), HellaSwag (10-shot), and HumanEval (0-shot), with accuracy serving as the primary evaluation proxy (higher is better). The experiment compares various quantization methods, including CALM, to comprehensively reflect the ability of each quantization method to maintain model performance across diverse tasks.

Table ~\ref{tab:downstream_split} presents a quantitative performance comparison of different models on multiple downstream tasks. Generally, all quantization methods result in accuracy degradation compared to the FP16 baseline; however, the magnitude of this decline varies significantly depending on the task type, model scale, and quantization methods. Notably, our proposed method (CALM) maintains optimal or near-optimal performance in most scenarios.

Distinct tasks exhibit varying sensitivities to quantization errors. Mathematical reasoning (GSM8K) and code generation (HumanEval) are the most fragile, showing the steepest accuracy drops across all models. For instance, on Llama-3-8B, accuracy reduction on GSM8K ranges from 2.84 (CALM) to 5.28 (GPTQ), while on HumanEval, the drop ranges from 1.04 (CALM) to 2.26 (GPTQ). In contrast, commonsense reasoning (HellaSwag) and multi-task understanding (MMLU) exhibit relatively minor degradation.

The efficacy of CALM extends to smaller, more sensitive models. On Qwen1.5-0.5B, CALM achieves a GSM8K score of 25.12, marking a substantial improvement of 3.25 and 4.90 over AWQ and GPTQ, respectively.

These empirical results strongly validate the effectiveness of the layer-wise adaptive hybrid quantization methods employed by CALM. The CALM framework effectively identifies and accommodates the heterogeneous quantization method requirements across different model layers, thereby achieving efficient, low-loss quantization while strictly preserving the model's core inferential capability.
\begin{table*}[ht]
\centering
\scriptsize 
\renewcommand{\arraystretch}{1.25} 
\setlength{\tabcolsep}{4pt} 
\begin{tabular}{lcccccccccccc} 
\toprule
\multirow{2}{*}{\textbf{Method}} & \multicolumn{4}{c}{\cellcolor{gray!10}\textbf{Llama-3-8B}} & \multicolumn{4}{c}{\cellcolor{gray!10}\textbf{Llama-3.2-3B}} & \multicolumn{4}{c}{\cellcolor{gray!10}\textbf{Llama-3.2-1B}} \\
\cmidrule(lr){2-5} \cmidrule(lr){6-9} \cmidrule(lr){10-13} 
 & \textbf{GSM} & \textbf{HE} & \textbf{HS} & \textbf{MMLU} & \textbf{GSM} & \textbf{HE} & \textbf{HS} & \textbf{MMLU} & \textbf{GSM} & \textbf{HE} & \textbf{HS} & \textbf{MMLU} \\
\midrule
FP16        & 77.17 & 60.71 & 77.39 & 67.31 & 68.86 & 48.80 & 70.05 & 61.44 & 36.44 & 32.14 & 59.94 & 46.38 \\
AWQ         & 72.94 & 58.92 & 76.78 & 64.41 & 57.65 & 46.69 & 68.97 & 58.69 & 26.44 & 22.32 & 56.78 & 40.23 \\
GPTQ        & 71.89 & 58.45 & 76.91 & 60.54 & 55.75 & 46.23 & 68.35 & 57.16 & 24.24 & 21.43 & 57.04 & 40.58 \\
SmoothQuant & 72.55 & 58.95 & 76.83 & 64.55 & 56.33 & 45.97 & 69.34 & 58.73 & 25.87 & 22.64 & 56.89 & 41.32 \\
SpinQuant   & 73.56 & 59.24 & \cellcolor{blue!10}\textbf{77.23} & 64.98 & 57.93 & 46.12 & \cellcolor{blue!10}\textbf{69.83} & 58.31 & 26.13 & 22.75 & 57.68 & 42.64 \\
\textbf{CALM} & \cellcolor{blue!10}\textbf{74.33} & \cellcolor{blue!10}\textbf{59.67} & 77.13 & \cellcolor{blue!10}\textbf{65.87} & \cellcolor{blue!10}\textbf{60.22} & \cellcolor{blue!10}\textbf{47.02} & 69.78 & \cellcolor{blue!10}\textbf{59.27} & \cellcolor{blue!10}\textbf{27.44} & \cellcolor{blue!10}\textbf{23.21} & \cellcolor{blue!10}\textbf{58.53} & \cellcolor{blue!10}\textbf{43.50} \\
\bottomrule
\end{tabular}
\vspace{1em} 
\begin{tabular}{lcccccccc} 
\toprule
\multirow{2}{*}{\textbf{Method}} & \multicolumn{4}{c}{\cellcolor{gray!10}\textbf{Qwen1.5-1.5B}} & \multicolumn{4}{c}{\cellcolor{gray!10}\textbf{Qwen1.5-0.5B}} \\
\cmidrule(lr){2-5} \cmidrule(lr){6-9}
 & \textbf{GSM} & \textbf{HE} & \textbf{HS} & \textbf{MMLU} & \textbf{GSM} & \textbf{HE} & \textbf{HS} & \textbf{MMLU} \\
\midrule
FP16        & 58.03 & 35.11 & 65.20 & 59.19 & 33.18 & 26.78 & 49.65 & 45.76 \\
AWQ         & 48.32 & 26.31 & 64.12 & 57.32 & 21.87 & 18.76 & 48.12 & 43.26 \\
GPTQ        & 47.12 & 24.40 & 63.73 & 56.91 & 20.22 & 19.04 & 47.38 & 42.82 \\
SmoothQuant & 48.64 & 25.97 & 63.98 & 57.69 & 22.34 & 19.84 & 47.69 & 43.67 \\
SpinQuant   & 49.01 & 26.87 & \cellcolor{blue!10}\textbf{64.87} & 57.84 & 23.45 & \cellcolor{blue!10}\textbf{20.64} & 47.98 & 43.44 \\
\textbf{CALM} & \cellcolor{blue!10}\textbf{51.36} & \cellcolor{blue!10}\textbf{29.34} & 64.56 & \cellcolor{blue!10}\textbf{58.11} & \cellcolor{blue!10}\textbf{25.12} & 20.04 & \cellcolor{blue!10}\textbf{48.68} & \cellcolor{blue!10}\textbf{44.12} \\
\bottomrule
\end{tabular}
\caption{Downstream task performance across five LLMs. Higher accuracy(↑) indicates better performance. Upper: Llama series; Lower: Qwen series. (GSM=GSM8K, HE=HumanEval,HS=HellaSwag).
}
\label{tab:downstream_split}
\end{table*}

\subsection{Configuration Experiments}

To rigorously evaluate the individual contributions of each component and validate the design rationale underlying the CALM framework, we conducted extensive configuration studies. 

\subsubsection{Impact of PTQ Candidate Diversity}
A central motivation of CALM is the premise that a single quantization method cannot optimally accommodate the distinct characteristics of every layer. To validate this, we systematically excluded individual quantization algorithms from the candidate pool and observed the resulting impact on model performance. As shown in table \ref{ablation_combined},experimental results demonstrate that retaining the full spectrum of candidate quantization methods yields superior performance across all evaluated models and tasks. The exclusion of any single algorithm invariably leads to a marked degradation in model fidelity.
\begin{table}[H]
\centering
\scriptsize 
\renewcommand{\arraystretch}{1.3} 
\setlength{\tabcolsep}{3pt}       
\begin{tabular}{lcccccc}
\toprule
\textbf{Variant} & \textbf{C4}$\downarrow$ & \textbf{W2}$\downarrow$ & \textbf{GSM}$\uparrow$ & \textbf{HE}$\uparrow$ & \textbf{HS}$\uparrow$ & \textbf{MMLU}$\uparrow$ \\
\midrule
\multicolumn{7}{c}{\cellcolor{gray!10}\textbf{Llama-3-8B}} \\
\midrule
\rowcolor{blue!10} 
\textbf{CALM (Full)} & \textbf{12.72} & \textbf{6.89} & \textbf{74.33} & \textbf{59.67} & \textbf{77.13} & \textbf{65.87} \\
w/o SQ & 13.55 & 7.65 & 73.12 & 58.80 & 76.50 & 64.60 \\
w/o AWQ & 12.95 & 7.08 & 72.85 & 58.45 & 76.85 & 64.20 \\
w/o GPTQ & 12.80 & 6.96 & 73.98 & 59.40 & 77.05 & 65.50 \\
\multicolumn{7}{c}{\cellcolor{gray!10}\textbf{Llama-3.2-3B}} \\
\midrule
\rowcolor{blue!10}
\textbf{CALM (Full)} & \textbf{17.41} & \textbf{9.63} & \textbf{60.22} & \textbf{47.02} & \textbf{69.78} & \textbf{59.27} \\
w/o SQ & 18.50 & 10.15 & 58.50 & 46.10 & 68.80 & 58.10 \\
w/o AWQ & 17.80 & 9.85 & 57.10 & 45.85 & 69.10 & 57.90 \\
w/o GPTQ & 17.55 & 9.70 & 59.65 & 46.80 & 69.60 & 59.05 \\
\multicolumn{7}{c}{\cellcolor{gray!10}\textbf{Llama-3.2-1B}} \\
\midrule
\rowcolor{blue!10}
\textbf{CALM (Full)} & \textbf{22.88} & \textbf{11.70} & \textbf{27.44} & \textbf{23.21} & \textbf{58.53} & \textbf{43.50} \\
w/o SQ & 23.90 & 12.35 & 26.55 & 22.40 & 57.10 & 41.80 \\
w/o AWQ & 23.20 & 11.92 & 26.21 & 22.10 & 57.90 & 41.50 \\
w/o GPTQ & 23.05 & 11.88 & 27.10 & 23.05 & 58.30 & 43.10 \\
\multicolumn{7}{c}{\cellcolor{gray!10}\textbf{Qwen1.5-1.5B}} \\
\midrule
\rowcolor{blue!10}
\textbf{CALM (Full)} & \textbf{16.59} & \textbf{12.87} & \textbf{51.36} & \textbf{29.34} & \textbf{64.56} & \textbf{58.11} \\
w/o SQ & 17.45 & 14.20 & 49.50 & 27.50 & 63.20 & 56.80 \\
w/o AWQ & 16.90 & 13.45 & 48.95 & 27.10 & 64.05 & 56.50 \\
w/o GPTQ & 16.70 & 13.05 & 50.90 & 29.05 & 64.40 & 57.85 \\
\multicolumn{7}{c}{\cellcolor{gray!10}\textbf{Qwen1.5-0.5B}} \\
\midrule
\rowcolor{blue!10}
\textbf{CALM (Full)} & \textbf{24.87} & \textbf{14.87} & \textbf{25.12} & \textbf{20.04} & \textbf{48.68} & \textbf{44.12} \\
w/o SQ & 26.10 & 16.50 & 22.50 & 18.90 & 47.50 & 42.90 \\
w/o AWQ & 25.40 & 15.50 & 23.10 & 19.20 & 48.10 & 43.10 \\
w/o GPTQ & 25.05 & 15.10 & 24.50 & 19.80 & 48.50 & 43.80 \\
\bottomrule
\end{tabular}
\caption{Downstream task performance across five LLMs. Higher accuracy($\uparrow$) indicates better performance.(GSM=GSM8K, HE=HumanEval, HS=HellaSwag). 'w/o' stands for 'without'.}
\label{ablation_combined}
\end{table}
For instance, on the Llama-3-8B model, our comprehensive strategy pool achieves a C4 PPL of 12.72. In contrast, removing SmoothQuant causes the C4 PPL to deteriorate significantly to 13.55, while excluding AWQ results in a drop in GSM8K reasoning accuracy from 74.33\% to 72.85\%.

These findings strongly attest to the indispensable complementary strengths of different algorithms in handling layer-specific feature distributions. Blindly excising any method diminishes the model's overall expressive capacity. It is only by synergistically integrating these diverse algorithms—allowing each to govern the specific layers best suited to its intrinsic properties—that we can transcend the performance ceilings of monolithic approaches. 

\subsubsection{Method-Heterogeneity and Bit-Heterogeneity}
While traditional mixed-precision methods optimize storage by assigning lower bit-widths (e.g., 2-bit) to less sensitive layers, we contend that this 'Bit-Heterogeneity' is suboptimal under low-bit constraints due to the severe reduction in representational capacity. To test this, we compared Bit-Heterogeneity (HAWQ-V2 \citep{HAWQ-V2} and SpQR \citep{Dettmers2024SpQR}) against CALM under an identical 4-bit average budget. Furthermore, we measured the real-world inference latency (ms/token) on a single NVIDIA A800 GPU, using a sequence length of 2048 to evaluate practical hardware efficiency. As shown in table~\ref{tab:Method-Heterogeneity}, on Llama-3-8B, the traditional mixed-precision baseline (forcing some layers to 2-bit) resulted in a Wiki2 PPL of 7.95.
\begin{table}[H]
\centering
\scriptsize
\renewcommand{\arraystretch}{1.25}
\setlength{\tabcolsep}{3.5pt}
\resizebox{\linewidth}{!}{
\begin{tabular}{l ccc}
\toprule
\textbf{Method} & \textbf{Wiki2} $\downarrow$ & \textbf{GSM8K} $\uparrow$ & \textbf{Latency (ms/token)} $\downarrow$ \\
\midrule
\multicolumn{4}{c}{\cellcolor{gray!10}\textbf{Llama-3-8B (W4A8, Seq=2048)}} \\
\midrule
HAWQ-V2 & 7.25 & 73.70 & 5.72 \\
SpQR    & 6.97 & 74.22 & 8.16 \\
\rowcolor{blue!10}
\textbf{CALM} & \textbf{6.89} & \textbf{74.33} & \textbf{4.38} \\
\midrule
\multicolumn{4}{c}{\cellcolor{gray!10}\textbf{Qwen1.5-1.5B (W4A8, Seq=2048)}} \\
\midrule
HAWQ-V2 & 13.23 & 49.83 & 1.58 \\
SpQR    & 12.92 & 51.16 & 2.24 \\
\rowcolor{blue!10}
\textbf{CALM} & \textbf{12.87} & \textbf{51.36} & \textbf{1.15} \\
\bottomrule
\end{tabular}
}
\caption{Comparison between Bit-Heterogeneity and CALM under the same average 4-bit constraint.}
\label{tab:Method-Heterogeneity}
\end{table}

CALM maintains 4-bit precision globally but varies the algorithm—applying SmoothQuant for activation-heavy layers and AWQ for weight-sensitive ones. CALM's superior results prove that the same algorithm cannot cope with the significant variance in layer-wise feature distributions. Instead of sacrificing bit-width, optimizing the algorithmic fit for each layer yields a far better trade-off between efficiency and accuracy.

\subsubsection{Performance at Lower Bit-widths}
\begin{table}[H]
\centering
\scriptsize 
\renewcommand{\arraystretch}{1.3} 
\setlength{\tabcolsep}{3pt}        
\begin{tabular}{lcccccc}
\toprule
\textbf{Method} & \textbf{C4}$\downarrow$ & \textbf{W2}$\downarrow$ & \textbf{GSM}$\uparrow$ & \textbf{HE}$\uparrow$ & \textbf{HS}$\uparrow$ & \textbf{MMLU}$\uparrow$ \\
\midrule
\multicolumn{7}{c}{\cellcolor{gray!10}\textbf{Llama-3-8B (3-bit)}} \\
\midrule
FP16  & 12.28 & 6.41 & 77.00 & 62.40 & 78.50 & 66.50 \\
SpinQuant & 14.40 & 8.55 & 57.00 & 46.20 & 72.10 & 58.50 \\
SmoothQuant & 15.10 & 8.60 & 56.50 & 45.80 & 71.80 & 57.20 \\
AWQ & 15.80 & 10.20 & 55.00 & 42.10 & 69.50 & 54.20 \\
GPTQ & 19.25 & 13.00 & 53.00 & 35.50 & 65.20 & 45.00 \\
\rowcolor{blue!10}
\textbf{CALM} & \textbf{13.55} & \textbf{7.60} & \textbf{61.50} & \textbf{50.50} & \textbf{74.50} & \textbf{60.20} \\ \midrule
\multicolumn{7}{c}{\cellcolor{gray!10}\textbf{Llama-3.2-3B (3-bit)}} \\
\midrule
FP16  & 17.24 & 9.53 & 77.70 & 49.50 & 70.20 & 63.40 \\
SpinQuant & 20.20 & 11.40 & 62.50 & 34.20 & 64.10 & 55.00 \\
SmoothQuant & 20.80 & 11.85 & 60.20 & 32.50 & 63.50 & 53.80 \\
AWQ & 21.50 & 12.50 & 52.00 & 28.40 & 60.20 & 50.50 \\
GPTQ & 27.22 & 20.34 & 45.00 & 20.50 & 55.80 & 45.00 \\
\rowcolor{blue!10}
\textbf{CALM} & \textbf{18.45} & \textbf{10.65} & \textbf{68.00} & \textbf{38.00} & \textbf{66.50} & \textbf{57.50} \\ 
\bottomrule
\end{tabular}
\caption{Comparison of predicted 3-bit performance across different quantization methods.(GSM=GSM8K, HE=HumanEval, HS=HellaSwag)}
\label{3bit_comparison}
\end{table}
The traditional PTQ method often experiences a non-linear collapse in accuracy when the bit width is reduced to 3 bits, due to its inability to handle outliers in the weight distribution. In contrast, CALM achieves a smooth degradation in performance through a collaborative optimization strategy, demonstrating the algorithm's adaptability to extremely low-bit-width environments. As shown in Table \ref{3bit_comparison}, experimental results demonstrate that CALM maintains high performance even under the extreme constraints of 3-bit quantization. 

\subsubsection{Fine-Grained Hyperparameter Tuning }
\begin{table}[H]
\centering
\scriptsize
\renewcommand{\arraystretch}{1.25}
\setlength{\tabcolsep}{3pt}
\resizebox{\linewidth}{!}{
\begin{tabular}{l cccccc}
\toprule
\textbf{Method} & \textbf{C4} $\downarrow$ & \textbf{Wiki2} $\downarrow$ & \textbf{GSM8K} $\uparrow$ & \textbf{HumanEval} $\uparrow$ & \textbf{HellaSwag} $\uparrow$ & \textbf{MMLU} $\uparrow$ \\
\midrule
\multicolumn{7}{c}{\cellcolor{gray!10}\textbf{Llama-3-8B}} \\
\midrule
CALM  & 12.72 & 6.89 & 74.33 & 59.67 & 77.13 & 65.87 \\
\rowcolor{blue!10}
\textbf{CALM+} & \textbf{12.61} & \textbf{6.72} & \textbf{75.35} & \textbf{60.05} & \textbf{77.26} & \textbf{66.32} \\
\midrule
\multicolumn{7}{c}{\cellcolor{gray!10}\textbf{Llama-3.2-1B}} \\
\midrule
CALM & 11.70 & 22.88 & 27.44 & 23.21 & 58.53 & 43.50 \\
\rowcolor{blue!10}
\textbf{CALM+} & \textbf{11.58} & \textbf{22.45} & \textbf{28.52} & \textbf{24.15} & \textbf{58.90} & \textbf{43.95} \\
\midrule
\multicolumn{7}{c}{\cellcolor{gray!10}\textbf{Qwen1.5-1.5B}} \\
\midrule
CALM  & 16.59 & 12.87 & 51.36 & 29.34 & 64.56 & 58.11 \\
\rowcolor{blue!10}
\textbf{CALM+} & \textbf{16.48} & \textbf{12.70} & \textbf{52.15} & \textbf{29.90} & \textbf{64.82} & \textbf{58.45} \\
\bottomrule
\end{tabular}
}
\caption{By expanding the search space to include hyperparameter variants, CALM+ outperforms the standard methods in all models.}
\label{tab:fine_grained_calm}
\end{table}

By extending the search space to joint algorithm-parameter variants ($\text{Pool}^+ = \{ \text{GPTQ}, \text{AWQ}, \text{SmoothQuant}_{\alpha \in \{0.3,0.4,0.5,0.6, 0.7\}} \}$), CALM+ precisely targets layer-specific sensitivities. $\alpha$ governs the migration strength of quantization difficulty via the smoothing scale $s$:
\begin{equation}
s = \frac{\max(|X|)^\alpha}{\max(|W|)^{1-\alpha}}
\label{eq:smooth_scale}
\end{equation}
This granular optimization delivers vital performance gains (Llama-3.2-1B: +1.08\% GSM8K) while further pushing the precision envelope of larger models, as detailed in Table \ref{tab:fine_grained_calm}. By adaptively tuning $\alpha$ layer-by-layer, CALM+ effectively mitigates this sensitivity, proving that parametric heterogeneity is a critical dimension for pushing the limits of low-bit quantization.

\section{conclusion}
We presented a CKA-guided modular quantization framework that enables layer-wise selection of heterogeneous PTQ algorithms under fixed low-bit constraints. Our study reveals that different transformer layers respond unevenly to distinct quantization methods, and leveraging this algorithmic diversity is crucial for maintaining functional fidelity. By using CKA as a reliable proxy for layer-level representation alignment, CALM constructs heterogeneously quantized LLMs without any retraining. These findings highlight that method heterogeneity, rather than bit-width heterogeneity alone, is a key factor for advancing post-training quantization of LLMs.

\section{limitations}
Deviation between Local Optima and Global Performance: The framework adopts a greedy layer-wise selection strategy aimed at approximating the global optimum through local optimization. While computationally feasible, this approach may fail to capture long-range inter-layer dependencies, potentially settling at a sub-optimal configuration in certain complex models.
Increased Offline Calibration Cost: Compared to single-method quantization, the search process of CALM involves evaluating multiple candidate algorithms for each layer. Consequently, the time required for offline calibration scales linearly with the number of algorithms in the candidate pool.
\bibliography{acl}

\appendix
\section{Appendix}
\label{sec:appendix}
\subsection{code}
\url{https://anonymous.4open.science/r/CALM-ACL}
\subsection{CKA}
\label{CKA}
\subsubsection{Mathematical Definition}
As introduced in Section 3, we utilize Linear CKA as the proxy for functional fidelity. Here, we provide the complete mathematical formulation regarding the centering process.

Formally, let $\mathbf{X} \in \mathbb{R}^{n \times d}$ denote the output activation matrix of the full-precision baseline and $\mathbf{Y} \in \mathbb{R}^{n \times d}$ be the activation from a candidate quantization method, where $n$ is the batch size and $d$ is the feature dimension. To quantitatively assess the structural alignment between feature representations invariant to orthogonal transformation and isotropic scaling, we employ Linear Centered Kernel Alignment (CKA).

We first introduce the centering matrix $\mathbf{H} = \mathbf{I}_n - \frac{1}{n}\mathbf{1}\mathbf{1}^\top$, where $\mathbf{I}_n$ is the identity matrix and $\mathbf{1}$ is a column vector of ones. Let $\tilde{\mathbf{X}} = \mathbf{H}\mathbf{X}$ and $\tilde{\mathbf{Y}} = \mathbf{H}\mathbf{Y}$ represent the centered feature matrices. The Linear CKA similarity score is defined as:

\begin{equation}
\text{CKA}(\mathbf{X}, \mathbf{Y}) = \frac{\|\tilde{\mathbf{Y}}^\top \tilde{\mathbf{X}}\|_F^2}{\|\tilde{\mathbf{X}}^\top \tilde{\mathbf{X}}\|_F \|\tilde{\mathbf{Y}}^\top \tilde{\mathbf{Y}}\|_F}
\end{equation}

where $\|\cdot\|_F$ denotes the Frobenius norm. In our experiments, the feature dimension $d$ corresponds to the output dimension of the FFN. For the Llama architecture, this is $d = 4 \times d_{\text{model}}$, where $d_{\text{model}}$ is the hidden size of the transformer block.

Crucially, $\mathbf{X}$ and $\mathbf{Y}$ are computed using inputs from previous optimized layers to capture local fidelity and compensate for accumulated errors. We calculate CKA at the FFN output (pre-residual) to isolate the quantization impact on the core transformation, avoiding masking effects from skip connections.

\begin{figure*}[ht]
    \centering
    \includegraphics[width=0.8\textwidth]{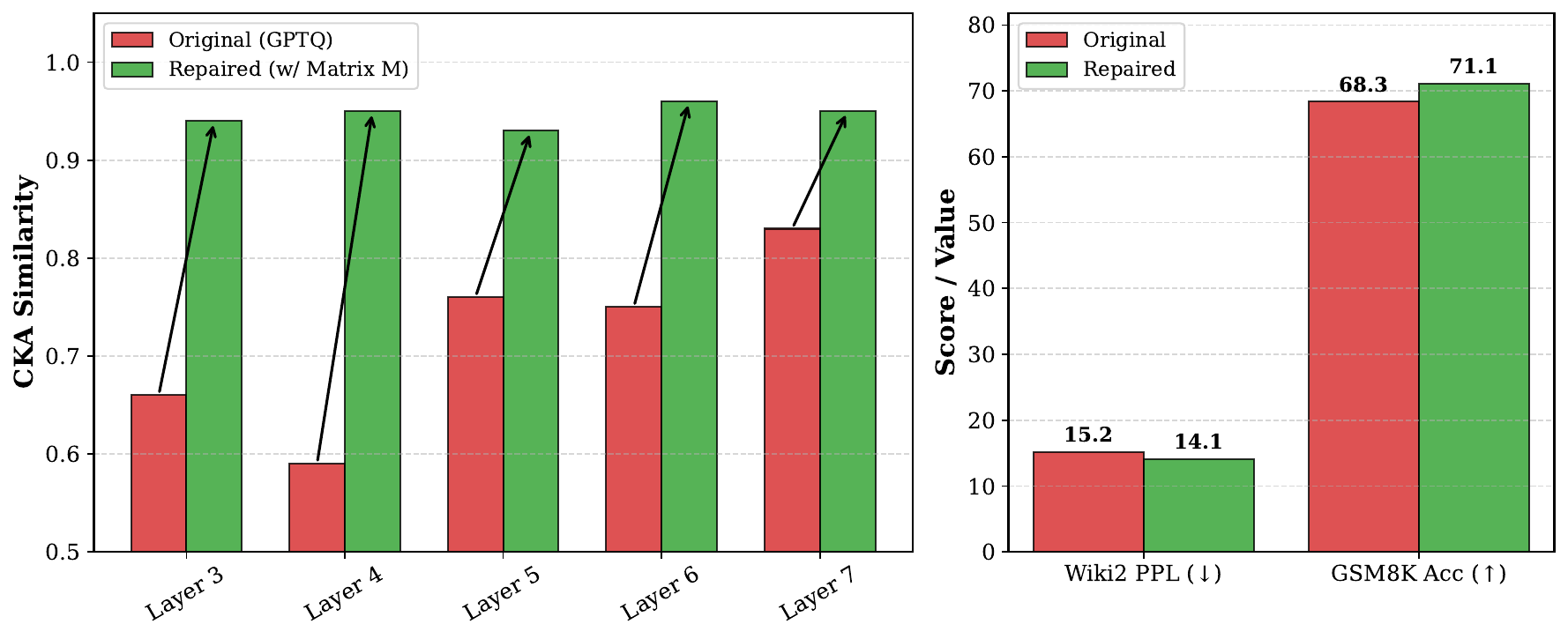} 
    \caption{Diagnostic experiment demonstrating that feature space restoration (via matrix M) recovers model performance}
    \label{repair}
\end{figure*}
\subsubsection{Theoretical Justification}
\label{cka_theory}
To rigorously validate the causal relationship between feature space alignment and model degradation, and to substantiate the efficacy of CKA as an optimization objective, we devised a diagnostic experiment based on linear reconstruction. We introduce a restoration protocol utilizing a Linear Transformation Matrix, $M$.Specifically, we identify the layer $l$ exhibiting the lowest CKA score under a homogeneous quantization method. 

Let $\mathbf{X}_{FP} \in \mathbb{R}^{B \times D}$ and $\mathbf{X}_{Q} \in \mathbb{R}^{B \times D}$ denote the activation outputs of the full-precision and quantized models for this layer, respectively. We formulate the following least-squares optimization objective to derive the optimal restoration matrix $\mathbf{M}$:
\begin{equation}
    \min_{\mathbf{M}} \| \mathbf{X}_{FP} - \mathbf{X}_{Q} \mathbf{M} \|_F^2
\end{equation}
By obtaining the closed-form solution to this linear regression problem, we absorb the resulting matrix $\mathbf{M}$ into the quantized weights of the layer (i.e., updating $\mathbf{W}'_{Q} \leftarrow \mathbf{W}_{Q} \cdot \mathbf{M}$), thereby achieving a targeted restoration of the feature manifold.

The results, illustrated in Figure \ref{repair}, are compelling. Upon applying matrix $M$, the CKA score of the target layer rebounds significantly, confirming that the feature space has been realigned to a near-FP16 state. Crucially, this local feature restoration translates into global performance gains: model perplexity (PPL) is markedly reduced, and accuracy on downstream tasks such as GSM8K sees substantial recovery.

This experiment provides critical empirical support for our hypothesis: a low CKA score accurately reflects structural disruption within the feature space. Consequently, maximizing feature alignment—whether by solving for matrix $M$ or selecting a superior quantization kernel—constitutes a robust mechanism for recovering model performance. These findings lay a solid theoretical foundation for the 'CKA-Guided Modular Quantization Framework' proposed in this work.

\subsection{Quantization Configuration}
\label{app:quant_config}

To ensure fair comparison and reproducibility, we adopt a unified quantization standard across all methods and ablation studies. Specifically, all models are quantized to 4-bit weights with 8-bit activations (W4A8) using a group size of 128. This rigorous standardization ensures that observed performance gains stem exclusively from algorithmic efficacy rather than hyperparameter variations.

Furthermore, we follow widely adopted practices by preserving sensitivity-critical components in full precision (FP16/BF16). These include all forward activations, the KV Cache, and the Embedding layers, as our preliminary analysis indicates that quantizing these modules leads to significant performance degradation. Within these constraints, our framework focuses on dynamically assigning the optimal quantization method to each intermediate Transformer layer.

\subsection{Impact of Quantization Granularity}
\begin{table}[H]
\centering
\scriptsize
\renewcommand{\arraystretch}{1.25}
\setlength{\tabcolsep}{3.5pt}
\resizebox{\linewidth}{!}{
\begin{tabular}{l cc cccc}
\toprule
\textbf{Method} & \textbf{C4} $\downarrow$ & \textbf{Wiki2} $\downarrow$ & \textbf{GSM8K} $\uparrow$ & \textbf{HumanEval} $\uparrow$ & \textbf{HellaSwag} $\uparrow$ & \textbf{MMLU} $\uparrow$ \\
\midrule
\multicolumn{7}{c}{\cellcolor{gray!10}\textbf{Llama-3-8B}} \\
\midrule
Block-8 & 13.15 & 7.20 & 73.05 & 58.75 & 76.65 & 64.70 \\
Block-4 & 12.89 & 7.05 & 73.65 & 59.10 & 76.90 & 65.10 \\
Block-2 & 12.78 & 6.94 & 74.10 & 59.45 & 77.05 & 65.60 \\
\rowcolor{blue!10}
\textbf{CALM} & \textbf{12.72} & \textbf{6.89} & \textbf{74.33} & \textbf{59.67} & \textbf{77.13} & \textbf{65.87} \\
\midrule
\multicolumn{7}{c}{\cellcolor{gray!10}\textbf{Qwen1.5-1.5B}} \\
\midrule
Block-8 & 17.25 & 13.90 & 49.10 & 27.20 & 63.45 & 56.90 \\
Block-4 & 16.95 & 13.40 & 50.07 & 28.10 & 63.90 & 57.35 \\
Block-2 & 16.70 & 13.05 & 50.95 & 28.90 & 64.35 & 57.80 \\
\rowcolor{blue!10}
\textbf{CALM} & \textbf{16.59} & \textbf{12.87} & \textbf{51.36} & \textbf{29.34} & \textbf{64.56} & \textbf{58.11} \\
\bottomrule
\end{tabular}
}
\caption{Performance comparison of varying quantization granularities.}
\label{tab:granularity_comparison}
\end{table}
To investigate the marginal effect of granularity on model fidelity, we compare our proposed layer-wise strategy against coarse-grained block-wise variants. We evaluate three baselines: Block-8, Block-4, and Block-2, where quantization configurations are shared across 8, 4, and 2 consecutive layers, respectively. Table \ref{tab:granularity_comparison} reveals that finer granularity consistently yields superior performance. Transitioning from Block-8 to Block-2 results in monotonic improvements across tasks. Crucially, CALM surpasses even the finest Block-2 baseline, validating the necessity of per-layer optimization. This superiority stems from the precise handling of sensitive layers characterized by outliers, preventing them from being forced into suboptimal methods inherent in shared block-wise configurations.

\subsection{Negligible Overhead}
\label{app:overhead}
\begin{table}[ht]
\centering
\scriptsize
\renewcommand{\arraystretch}{1.25}
\setlength{\tabcolsep}{3.5pt}
\resizebox{\linewidth}{!}{
\begin{tabular}{l ccc ccc}
\toprule
\textbf{Method} & \textbf{Fwd. (ms)} $\downarrow$ & \textbf{Gen. (T/s)} $\uparrow$ & \textbf{Fwd. Overhead} & \textbf{Gen. Overhead} \\
\midrule
\multicolumn{5}{c}{\cellcolor{gray!10}\textbf{Llama-3.2-1B (A800, W4A8)}} \\
\midrule
\rowcolor{blue!10}
GPTQ & 1.14 $\pm$ 0.05 & 875.2 & - & - \\
CALM & 1.15 $\pm$ 0.06 & 869.5 & +0.87\% & -0.65\% \\
\midrule
\multicolumn{5}{c}{\cellcolor{gray!10}\textbf{Llama-3.2-3B (A800, W4A8)}} \\
\midrule
\rowcolor{blue!10}
GPTQ & 2.50 $\pm$ 0.10 & 398.5 & - & - \\
CALM & 2.52 $\pm$ 0.12 & 396.1 & +0.80\% & -0.60\% \\
\midrule
\multicolumn{5}{c}{\cellcolor{gray!10}\textbf{Llama-3.1-8B (A800, W4A8)}} \\
\midrule
\rowcolor{blue!10}
GPTQ & 4.35 $\pm$ 0.15 & 229.8 & - & - \\
CALM & 4.38 $\pm$ 0.18 & 228.3 & +0.69\% & -0.65\% \\
\bottomrule
\end{tabular}
}
\caption{Inference latency and throughput comparison on a single NVIDIA A800 GPU using highly optimized kernels (Sequence Length = 2048). Overhead represents the relative change of CALM compared to the GPTQ baseline.}
\label{tab:latency_efficiency}
\end{table}
The efficiency of CALM is further validated through real-world latency measurements on an NVIDIA A800 GPU. As shown in Table \ref{tab:latency_efficiency}, compared to the standard uniform quantization method (GPTQ), CALM introduces negligible computational overhead. Across various model scales, the average latency increase is only $\sim$0.60\% for the forward pass and $\sim$0.07\% for token generation. This demonstrates that CALM provides significant accuracy improvements without sacrificing inference speed, making it highly practical for latency-sensitive applications.
\end{document}